\documentclass{article}

\usepackage[preprint]{neurips_2026}

\usepackage[utf8]{inputenc} %
\usepackage[T1]{fontenc}    %
\usepackage{hyperref}       %
\usepackage{url}            %
\usepackage{booktabs}       %
\usepackage{amsfonts}       %
\usepackage{nicefrac}       %
\usepackage{microtype}      %
\usepackage{xcolor}         %
\usepackage{xspace}
\usepackage{graphicx} 
\usepackage{multirow}
\usepackage[table]{xcolor}
\usepackage[most]{tcolorbox}
\usepackage{xcolor}
\usepackage{wrapfig}
\usepackage[normalem]{ulem}
\usepackage{float}
\usepackage{placeins}

\definecolor{linkblue}{RGB}{0, 70, 140}

\hypersetup{
  colorlinks=true,
  linkcolor=linkblue,
  citecolor=linkblue,
  urlcolor=linkblue
}

\title{\tech: Orchestrating Coding Agents for Open-Ended Discovery}

\author{%
Yuvraj Virk \quad
Zack Edds \quad
Chunqiu Steven Xia \quad
Lingming Zhang
\\[0.2em]
\uiuc{University of Illinois Urbana-Champaign}
\\[0.2em]
\texttt{\{yvirk2, zedds2, chunqiu2, lingming\}@illinois.edu}
}

\newcommand{\tech}{\textsc{SwarmResearch}\xspace}

\newcommand{\orchestrator}{Shepherd Agent}
\newcommand{\subagent}{Search Agent}

\newcommand{\llm}{LLM\xspace}

\newcommand{\Comment}[1]{}

\usepackage[most]{tcolorbox}
\newtcolorbox{resultbox}{
  colback=gray!5,
  colframe=black!60,
  boxrule=0.8pt,
  arc=2mm,
  left=6pt,
  right=6pt,
  top=6pt,
  bottom=6pt
}

\usepackage[most]{tcolorbox}
\usepackage{xcolor}

\newtcblisting{skillprompt}[1][]{
  listing only,
  breakable,
  enhanced,
  colback=gray!3,
  colframe=gray!45,
  boxrule=0.4pt,
  arc=2pt,
  left=6pt,
  right=6pt,
  top=6pt,
  bottom=6pt,
  listing options={
    basicstyle=\ttfamily\footnotesize,
    breaklines=true,
    breakatwhitespace=false,
    columns=fullflexible,
    keepspaces=true,
    showstringspaces=false,
    tabsize=2
  },
  #1
}

\newcommand{\uiuc}[1]{{#1\textsuperscript{\includegraphics[scale=0.004]{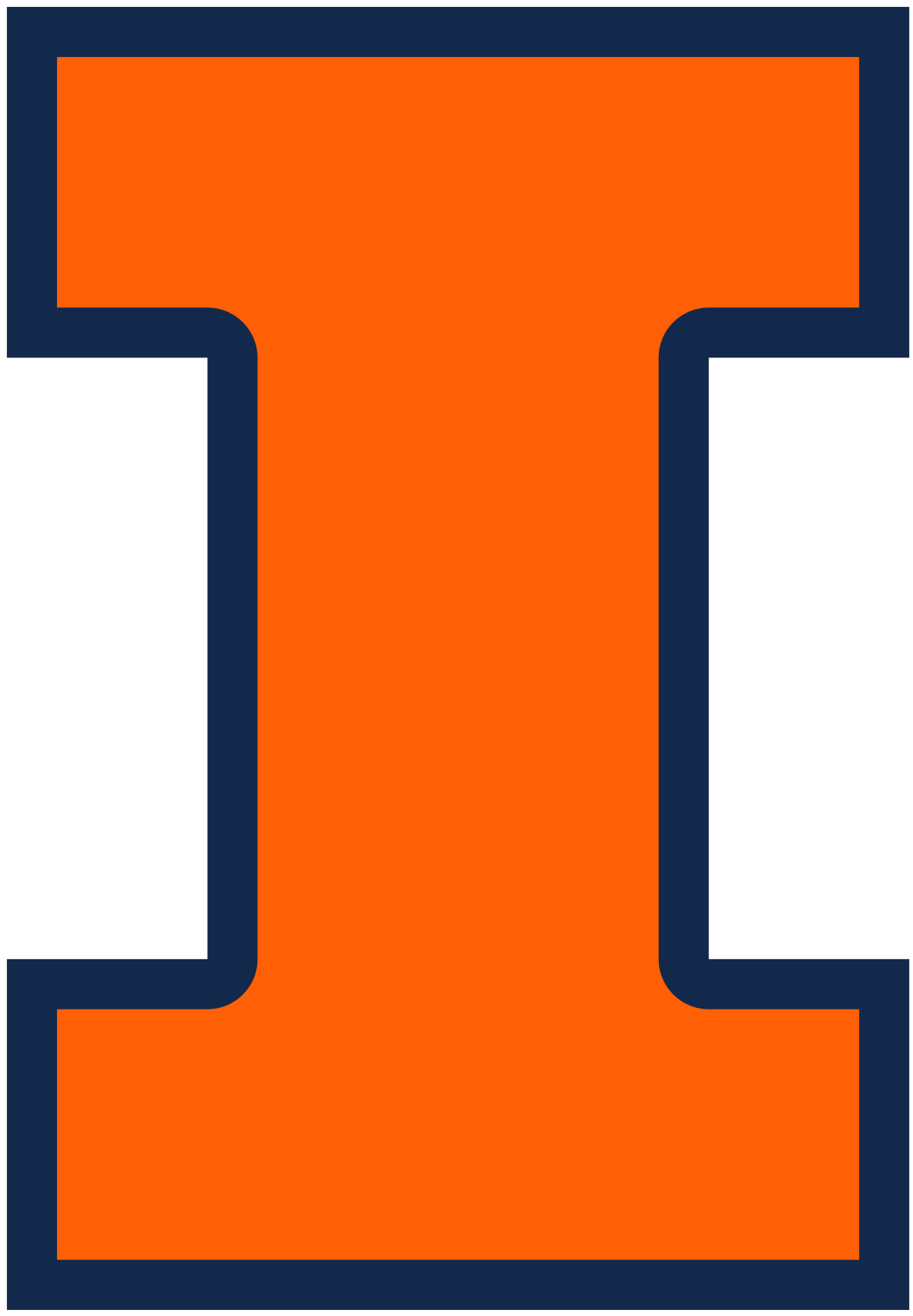}}}}

\begin{document}

\frenchspacing

\maketitle

\vspace{-1.5em}

\begin{abstract}
\vspace{-0.7em}
Long-running coding agents such as autoresearch can persistently discover optimizations for open-ended problems. However, they tend to converge onto a single high-level approach, then proceed with low-level edits while missing other superior approaches to the problem. We hypothesize two harness-level design choices contribute to this behavior: accumulating context in a single long-running agent and only exposing a single program state to edit. We introduce \tech, an orchestrator-subagent harness in which a Shepherd Agent uses \textit{global context} to steer a population of Search Agents, each operating with \textit{local context} in their respective git branch. On open-ended optimization tasks, \tech discovers better or comparable solutions to state-of-the-art LLM-guided evolution and multi-agent techniques on 13/15 tasks, driven by higher-level exploration. Compared with fixed scaling of serial and parallel agents, \tech's orchestrator-guided scaling discovers better-performing solutions by adapting parallelism at different search depths. Code and details available at \href{https://github.com/SwarmResearch/SwarmResearch}{\textcolor{linkblue}{https://github.com/SwarmResearch/SwarmResearch}}. 
\end{abstract}
\vspace{-0.5em}
\begin{figure}[h]
    \includegraphics[width=1.02\linewidth]{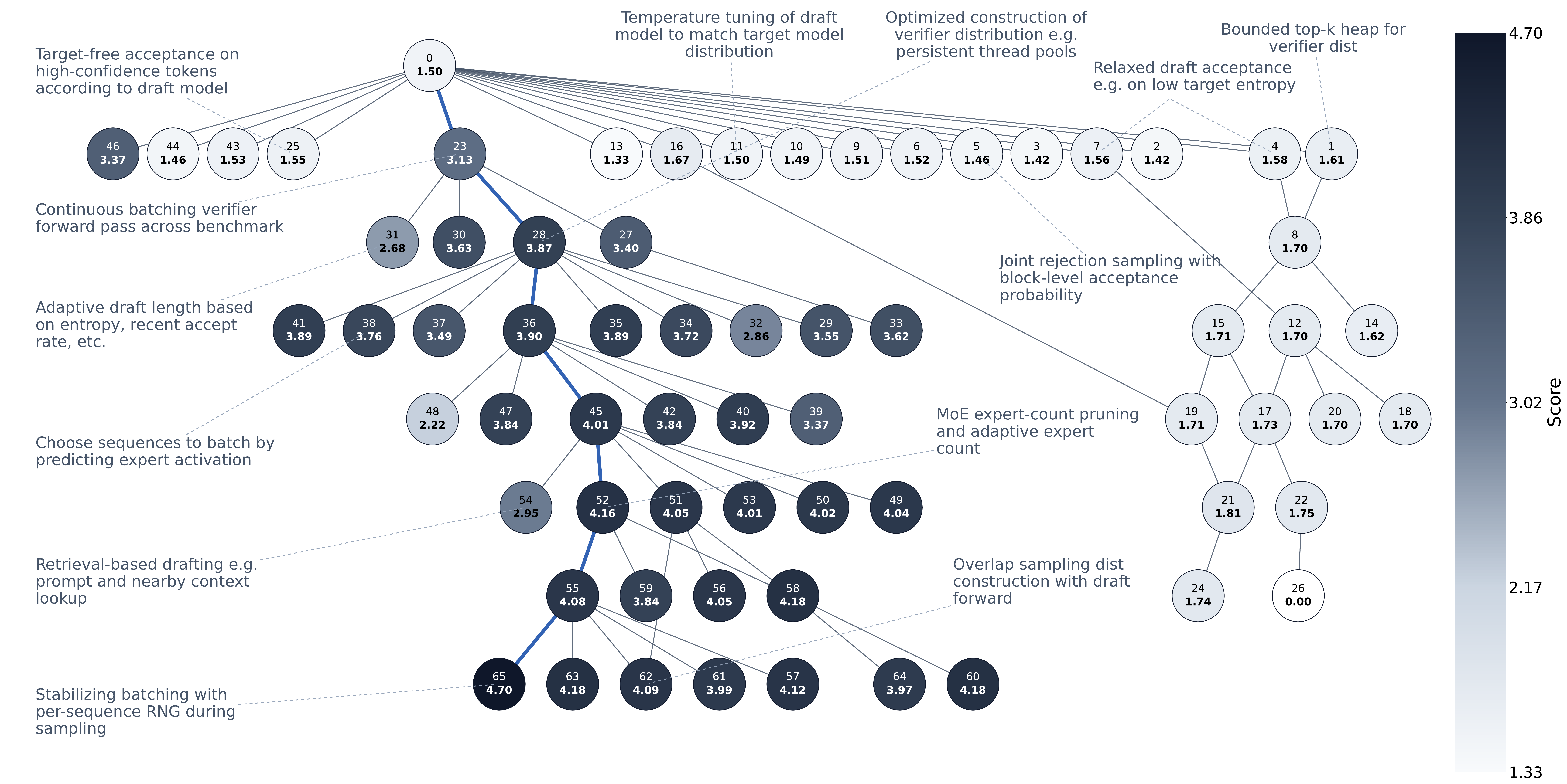} 
    \vspace{-1.2em}
    \caption{
    \tech designs speculative decoding implementations. Nodes represent spawned search agents and their solutions. Edges indicate the lower node builds on the solution above. The top node value shows spawn order; the bottom value shows speedup over vanilla decoding. \tech explores diverse approaches over $\sim$11 hours without converging onto any one approach. On 100 held-out reasoning-intensive tasks (averaged across 5 seeds), \textbf{\tech's achieved mean tok/s is 4.58$\times$ faster than naive vanilla decoding}. After $\sim$12 hours, \textbf{vanilla autoresearch’s}~\citep{autoresearch} \textbf{speedup is only 1.80$\times$} while \textbf{CORAL’s}~\citep{coral2026} \textbf{is 2.26$\times$}. All methods use Opus 4.8 Claude Code. Details are in Section~\ref{sec:spec-dec-case-study}. 
    }
    \label{fig:evo-graph}
    \vspace{-2em}
\end{figure}

\section{Introduction}
Open-ended research and engineering problems often have vast solution spaces and no known optimum. 
An example is designing fast decoding algorithms for LLM inference: researchers pursue many different directions, from KV-cache management ~\citep{kwon2023efficient} to improved speculative decoding schemes ~\citep{leviathan2023fast}. 
One way they evaluate progress is by end-to-end algorithm speed. 
These directions can be viewed as distinct regions of an optimization landscape, where some regions correspond to weak approaches which contain poor local optima, while others contain substantially faster solutions.
Making steady progress therefore requires continuous exploration of diverse, high-level approaches instead of immediately committing to a single neighborhood of the search space.

Test-time scaling~\citep{brown2024large} has shown promise in eliciting LLMs’ ability to search open-ended solution spaces. 
LLM-guided evolution iteratively samples an LLM to generate a new solution given select context on solutions generated so far; heuristic algorithms control exact context to guide search behavior~\citep{novikov2025alphaevolve, openevolve, liu2026evox, cemri2026adaevolve, lange2025shinka}. 
However, \textit{long-running coding agents} may support more adaptive and complex search by better expressing frontier model capabilities than heuristic LLM-guided evolution.
These agents can iteratively implement and test complex code, decide when to refine an existing approach versus pivot to a new direction, and use external tools such as web search. 
Autoresearch~\citep{autoresearch} is a popular example: it runs off-the-shelf coding agents such as Claude Code for hours or days in an experimental loop. One iteration implements changes, evaluates them, commits improvements, and reverts regressions. 
Autoresearch has inspired projects across domains with impressive results in systems optimization~\citep{lutke2026liquidpr2056}, kernel optimization~\citep{rightnow2026autokernel}, and prompt injection~\citep{panfilov2026claudini}.

Yet, long-running coding agents tend to converge quickly on a single high-level approach, then spend hours making small edits and micro-optimizations. 
Rather than exploring diverse directions, they can become trapped in a suboptimal region of the search space. 
We view this as a harness-engineering problem: context management and program versioning bottleneck divergent discovery. 
As agents accumulate long conversation histories filled with incremental refinements on one approach, substantial shifts in approach become unlikely.
By committing all improvements to a single editable program, coding agent experimental loops do not preserve alternative directions for further exploration, effectively turning the process into greedy local search. 

Instead, \tech uses an orchestrator to delegate experiment ideation and execution to subagents with fresh or controlled context windows, and it assigns all major edits to separate git branches. \tech is implemented as three skills compatible with popular coding agent tools such as Claude Code. A \textit{\orchestrator}\xspace spawns and steers a population of \textit{\subagent s}. Each \subagent\xspace proposes and executes new ideas given minimal context, while the \orchestrator\xspace maintains global context over all \subagent\xspace activity and strategically steers the population toward productive, diverse exploration.

We compare \tech on state-of-the-art LLM-guided evolution~\citep{liu2026evox} and multi-agent techniques~\citep{coral2026} on 15 open-ended optimization tasks across mathematics, heuristics, and competitive heuristic engineering. \tech exceeds or matches these techniques on 13/15 tasks. It typically makes 1.7$\times$ to 3.2$\times$ times larger code changes, reflecting higher-level exploration of approaches. We analyze \tech from a test-time scaling perspective. By using an orchestrator to autonomously vary the number of parallel agents at different search depths, it exceeds the performance of optimal fixed scaling of parallel and serial agents on 4/5 tasks. Finally, we present a detailed case study on speculative decoding to illustrate \tech's capabilities and limitations.

\begin{figure}[t]
    \centering
    \includegraphics[width=\linewidth]{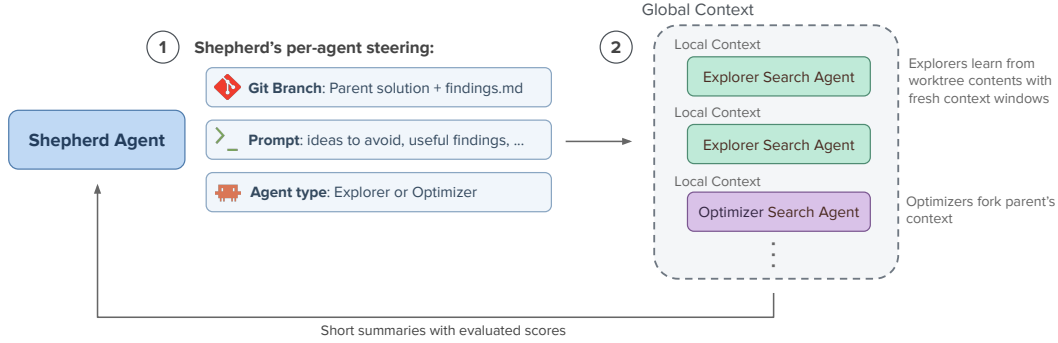}
    \caption{\textbf{Overview of \tech.} Implements an orchestrator-subagent loop: \textbf{(1)} \textit{Shepherd Agent} spawns concurrent search agents with 3 per-agent steering mechanisms: selecting a parent solution by setting up its git branch, writing a prompt to share minimal context, and selecting the Search Agent type. \textbf{(2)} \textit{Search Agents} ideate, implement, evaluate, and iterate on a solution. \textit{Explorer Search Agents} have fresh context windows so they're unanchored to prior work, while \textit{Optimizer Search Agents} fork their parent's conversation history to continue along its detailed history of attempts. The Shepherd Agent receives summaries back from all Search Agents to strategically steer new ones.
    }
    \label{fig:overview}
    \vspace{-1em}
\end{figure}

\section{Approach}

We propose \tech, an orchestrator-subagent harness in which a Shepherd Agent uses global context to steer a population of Search Agents, each operating in their respective git branch with fresh or controlled local context. Figure \ref{fig:overview} describes an overview of the agent architecture. Our design is centered around two problems in long-running coding agents that bottleneck divergent solution discovery: accumulating context around one approach and only editing one program state. 

First, early exploration by long-running coding agents often only covers a small number of high-level approaches. Once one approach appears promising, the agent proceeds with a long-tail of low-level edits and micro-optimizations on it. Experimenting with new high-level approaches requires several rounds of large program rewrites, which are unlikely when the agent is conditioned on a long history of incremental refinements to the current approach. As shown in Figure \ref{fig:overview}, to reduce this bias, the Shepherd Agent delegates experiment ideation and execution to Search Agents with fresh or controlled context windows.

Second, long-running coding agents edit a single program state: all improvements are committed, regressions are reverted, and future search proceeds from the current best implementation. This loop design makes maintaining competing high-level directions difficult and encourages greedy local search. 
Ideally, an agent harness should let search branch naturally, so that multiple different versions of a program can evolve independently when they instantiate different ideas. 
As shown in Figure \ref{fig:overview}, the Shepherd Agent assigns a new git branch for every Search Agent so all major edits occupy different versions. This way, promising yet not top-performing solutions are not prematurely discarded and can be refined. We describe the harness design, context management, and git management strategies in detail below. See Appendix~\ref{app:skills} for the full skills.

\subsection{Architecture}
The \textit{\orchestrator} is an orchestrator responsible for spawning and steering waves of concurrent search agents, while \textit{\subagent s} are subagents responsible for ideating a new approach, implementing, evaluating, and iterating on it. The \orchestrator \xspace is given a \subagent \xspace budget to manage. 
Unlike classic orchestrator-subagent architectures, which partition complex tasks such as large-scale code generation across subagents, the \orchestrator\xspace does not decompose the open-ended problem for handoff.
Instead, it steers population-level search behavior: our skills emphasize initializing and maintaining a diverse population of ideas, prioritizing search agent budget on promising ideas, and breaking out of plateaus. 
The \orchestrator\xspace accomplishes these goals using 3 steering mechanisms (Figure~\ref{fig:overview}; Step 1) when spawning new \subagent s: parent selection, search agent type, and prompts.

\textbf{Parent Selection.} For every new Search Agent, the Shepherd Agent creates a new git branch and worktree from either a task's baseline solution or a completed Search Agent’s commit. Since each Search Agent begins by reading and modifying this \textit{parent solution} in its worktree, that parent solution determines the neighborhood of solutions the agent explores. 
Parent selection lets the Shepherd Agent steer population search without prescribing specific ideas. Branches from the initial baseline will encourage new from-scratch exploration, while branches from completed agents will direct effort toward promising or underdeveloped directions.

In traditional evolutionary methods~\citep{novikov2025alphaevolve, openevolve, lange2025shinka}, parent selection is also used to direct exploration-exploitation behavior but is completed by a heuristic algorithm rather than an LLM. 
In existing single and multi-agent approaches~\citep{autoresearch, coral2026}, parent selection is implicit.
Agents \textit{independently} reflect on their conversation history and shared memory to decide which solution to expand. 
In contrast, \tech uses an orchestrator to distribute parents strategically rather than relying on  uncoordinated agent decisions.

\textbf{Search agent type.} \textit{Explorer Search Agents} have fresh context windows and use the content in their worktree as context. Their goal is to explore new approaches. \textit{Optimizer Search Agents} fork their parent agent's conversation history, in addition to using worktree content as context. Their goal is to make a few refinements to the parent solution. 

\textbf{Prompts.} Since Search Agents only have local context of their parent, the Shepherd Agent can supplement their context through prompts that share minimal, relevant context or add directions. For example, if Explorer Agents repeat similar ideas, prompts can provide a summary of already explored ideas and instruct them to explore new ideas. 
A prompt could also include complementary findings by other Search Agents. 
On plateaus, the prompt could include a briefing of strengths and weaknesses of the best approaches, with a suggestion to identify novel techniques targeting limitations. 
We found that restricting the Shepherd Agent from prescribing specific ideas in prompts helped maintain idea diversity; otherwise, the Shepherd Agent itself could get stuck inside a basin of ideas.

\subsection{Context Management}
\label{sec:context_management}

\begin{figure}[h!]
    \centering
    \includegraphics[width=0.8\linewidth]{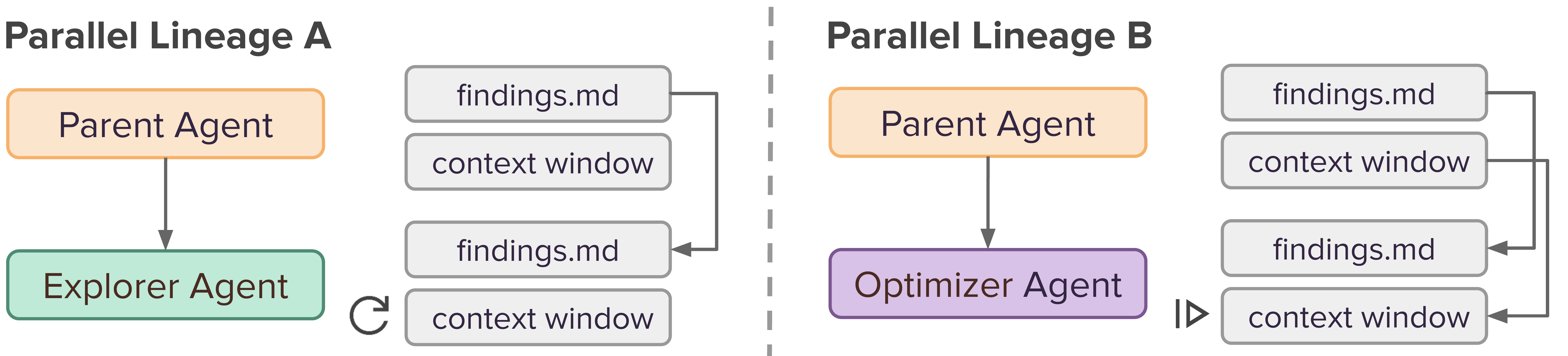}
    \caption{\tech context management for search agents.}
    \label{fig:context-management}
\end{figure}

\tech separates context into two tiers. Search Agents operate with \textit{local} context: the contents of their worktree and, for Optimizer Agents, the conversation history of their direct parent. The Shepherd Agent operates with \textit{global} context: summaries of all Search Agent attempts, including their approach and evaluated score. This separation is designed to preserve diversity among Search Agents while still giving the Shepherd Agent enough information to strategically steer the population.

A Search Agent's persistent state is primarily captured by its worktree. As shown in Figure~\ref{fig:context-management}, Explorer Agents start with fresh context windows, so their only context is the contents of the worktree they are assigned. 
Fresh context makes each agent less anchored to previous work and better suited for trying higher-level changes in approach.
Optimizer Agents inherit their parent's conversation history in addition to the worktree, giving them access to a detailed log of successes and failures to continue long serial refinement when an approach appears promising. Before committing their changes, each Search Agent appends a short summary of its approach and evaluation results to a \texttt{findings.md} file. It serves as a lineage-local log of attempts and outcomes, giving future Search Agents lightweight access to their ancestors' attempts without anchoring them in a long conversation history.

Our orchestrator-subagent design addresses two failure modes of context accumulation. 
First, standalone coding agents tend to accumulate long conversation histories that repeatedly refine a small number of high-level approaches. 
These histories can make larger pivots less likely, because the agent is repeatedly conditioned on the current approach and its incremental refinements. \tech avoids this diversity loss by spawning Explorer Agents with fresh context windows, while still supporting long-tail improvement through Optimizer Agents that inherit detailed conversation history when serial refinement is useful.

Second, multi-agent systems that self-organize through a shared memory e.g. CORAL~\citep{coral2026}, rather than an orchestrator, also converge onto a single high-level approach. When all agents see that another agent found a stronger solution, they decide to improve the best solution and abandon independent directions. \tech eliminates this possibility by restricting Search Agent context to their direct lineage, rather than exposing them to all lineages. 
At the same time, global information remains useful for strategy, by revealing which approaches are promising, which fall behind, and where the population lacks coverage. \tech incorporates this information through the Shepherd Agent, which uses global context to distribute Search Agents across parents and write prompts that selectively pass relevant information to new Search Agents.

\begin{wrapfigure}{r}{0.33\textwidth}
    \vspace{-1.5em}
    \centering
    \includegraphics[width=\linewidth]{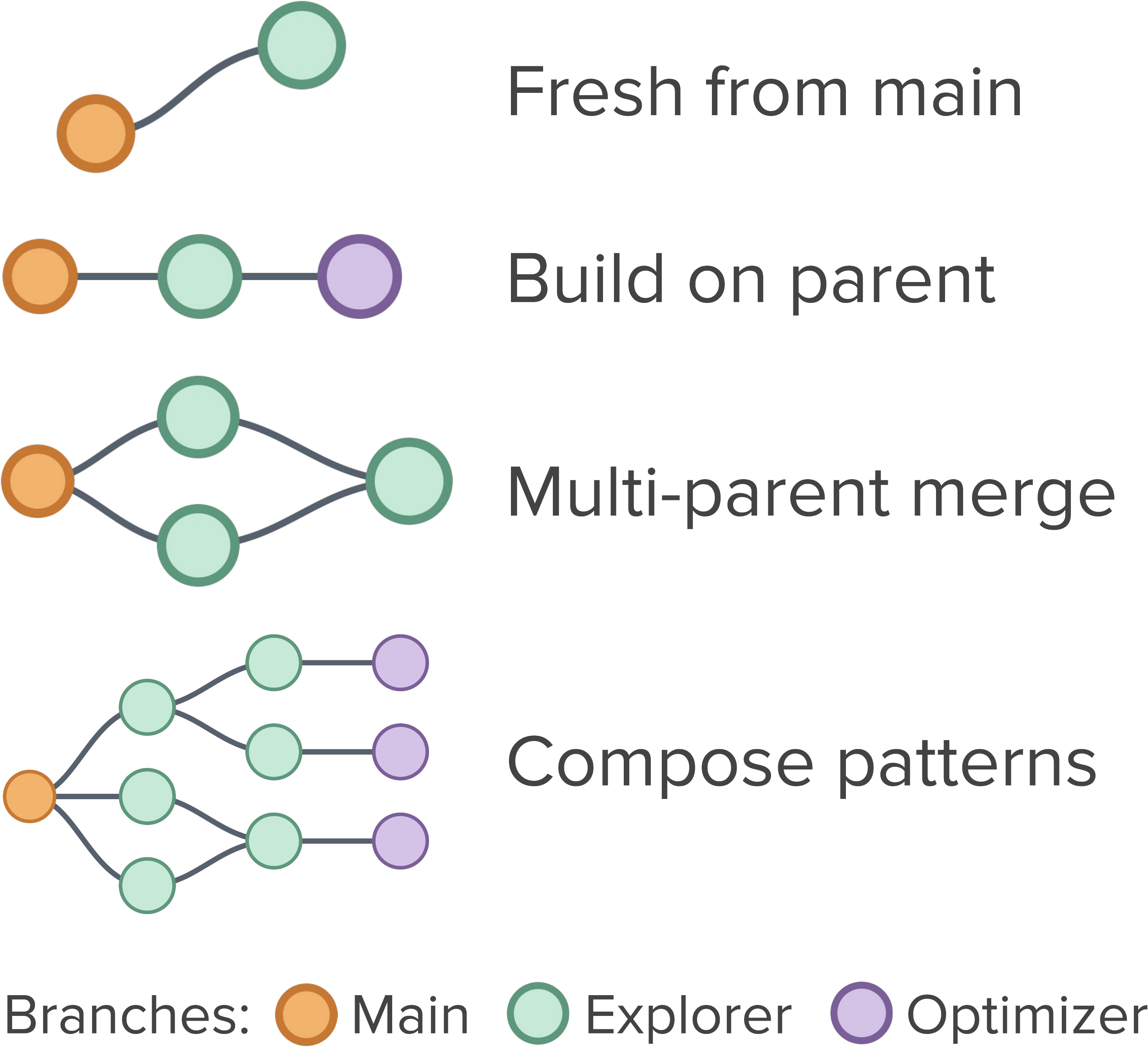}
    \caption{Git management for Search Agents.}
    \vspace{-1em}
    \label{fig:git-management}
\end{wrapfigure}

\subsection{Git Management}
\label{sec:git_management}

The Shepherd Agent assigns every Search Agent to a new git branch and worktree. 
The \texttt{main} branch contains only the initial task setup: a \texttt{prompt.md} file defining the problem and a minimal working baseline solution. As shown in Figure~\ref{fig:git-management}, branches can start from \texttt{main} to try a new from-scratch approach, from a completed Search Agent's commit to iterate on a prior solution, or merge multiple parent branches to combine different solutions. 
Branching allows the Shepherd Agent to express a wide range of search patterns. For example, it can spawn multiple concurrent agents from the same parent to fan out
around a promising solution, or spawn a serial chain of explorers and optimizers to build on a lineage's successes and failures. Since every Search Agent is versioned, fan-out and serial iteration can be flexibly interleaved across time and distant regions of the search space.

The branching approach in \tech contrasts with existing long-running coding agents, which usually make successive edits to a single program on a single branch. Since improvements are kept and regressions are reverted, the agent's search trajectory becomes a greedy sequence of local refinements. \tech 's branch-based versioning instead preserves competing approaches and avoids collapsing the search back into the current best implementation, making it easier to revisit, extend, combine, or abandon approaches.

\section{Evaluation}

\subsection{Experimental Setup}

\textbf{Benchmark.} 
Our benchmark consists of open-ended optimization problems with no known optimum. 
They fall in 3 domains: math, systems, and heuristic algorithms. 
Most mathematical optimization tasks require writing optimization programs to generate constructions demonstrating new bounds on a problem. We use 5 math tasks also used by prior work ~\citep{liu2026evox, coral2026, novikov2025alphaevolve}. Systems tasks are from the ADRS benchmark~\citep{cheng2025barbarians}, which is based on real-world systems research. Our math and systems tasks match the benchmark suite used by our multi-agent baseline CORAL~\citep{coral2026} (excluding MMD-16-2 for cost because of its similarity to MMD-14-3). Heuristics tasks are from ALE-Bench-Lite~\citep{imajuku2025alebench}, which pulls problems from the AtCoder Heuristics Contest. These competitive heuristic algorithm problems are inspired by optimization challenges from industry. Our 5 problem subset was selected by top-level solution categories that reward high-level algorithm exploration over tuning of generic optimization patterns. We use ALE-Bench's suggested performance metric which typically ranges from 0 to 3500. This metric is derived from the relative rankings of competition participants for a problem. All tasks include a minimal working baseline solution.

\textbf{Baselines and configurations.} We compare \tech to the top-performing evolutionary baseline EvoX~\citep{liu2026evox} and the state-of-the-art multi-agent system CORAL~\citep{coral2026}. EvoX adapts search behavior by generating parent selection programs on-the-fly using an LLM. In CORAL, concurrent coding agents run experimental loops and use a shared filesystem-based memory to exchange insights and build on one another’s successes and failures. Periodic prompts instruct agents to reflect, pivot on plateaus, and consolidate findings into memory. 
We configure CORAL to use 4 agents, as in their paper. 
Both CORAL and \tech are ran until a \$50 budget is exhausted. EvoX is executed for 100 iterations as configured in their paper, after which it has diminishing returns, with average task cost \$23.50. Both CORAL and Shepherd use Claude Code as the agent runtime. Note that CORAL and \tech's solution performances do not converge within the budget but reach a point of diminishing returns. CORAL settles on a top approach and subsequent gains would result from long-tail polishing. No approach has internet access. CORAL and \tech only have access to evaluators as APIs so their internals are hidden.

We note that EvoX achieves poorer performance with Opus 4.6 on some problems than presented in their original paper which used GPT-5. Small-scale experiments indicate that EvoX performs significantly better using GPT models, so its ranking may change if the latest GPT models are used. We only report Opus 4.6 results due to cost limitations and since model dependency is a real limitation of the approach. Additionally, our results obtained by reproducing CORAL with their publicly available implementation is worse than their originally reported results. However, we include their originally reported results as the state-of-the-art where applicable. Finally, we run every technique once per task. Although one run does not give a stable estimate of every technique's per-task performance, obtaining stable estimates are unrealistically expensive. Given realistic budget constraints and the stochastic nature of LLM-based discovery techniques, one long run per task achieves the most \textit{representative} results as per suggested use by each baseline. For a more detailed discussion of between-run variance, see Appendix~\ref{app:variance}.

\textbf{Granularity of Code Changes.} Low-level changes appear as targeted edits to a parent program. High-level changes in approach appear as rewrites of a parent program and thus associated with a higher number of lines of code changes. For \tech and CORAL, we compare the lines changed between every commit and its parent commit. EvoX only logs programs that improved over their parents, so we only take the changed lines of code for those programs and their parent. For each task, we compute the mean lines of code changed per attempt by a given method. Then, we compare the median of these task-level means.

\label{fixed_scaling_configs}
\textbf{Comparison to Optimal Fixed Scaling.} Fixed scaling is our naive test-time scaling baseline. By fixed scaling, we mean executing $n$ concurrent agents for $k$ serial iterations. An iteration consists of one-shot generating/editing a solution and evaluating it. On the other hand, \tech uses orchestrator-guided scaling, where an orchestrator dynamically decides when to launch a serial agent off a parent or a parallel agent off another parent. Does orchestrator-guided scaling exceed the performance of fixed scaling? Instead of completely independent agents, we find sharing a git history between them performs better. Shared git history enables parallel agents to switch to refining another agent's better solution. Here, an iteration consists of exploring the shared git history, generating a solution, evaluating it, and making a descriptive commit back to the shared git history. This way, separate agents can switch to better approaches discovered by other agents. Concurrent agents are synchronized at each iteration so they make consistent use of the shared git history. Fixed scaling is tested using the minimal coding agent harness Pi~\citep{zechner2025picodingagent} with Minimax-M2.5. We test 5 configurations of $(n,k)$ within a 60 total iteration budget: (5, 12), (10, 6), (15, 4), (20, 3), (30, 2). Results are averaged over 3 runs and shown for the best-performing configuration. 

To isolate the effect of using an orchestrator-subagent architecture, we use a minimal \tech harness in Pi. The orchestrator has a subagent tool that only accepts a \textit{branch} field for parent selection. The orchestrator can not write prompts for subagents and Explorer Agents are the only subagent type. Spawning a subagent runs one serial iteration off the parent without access to a shared git history. Subagents use Minimax-M2.5 while the orchestrator uses Claude Sonnet-4.6; Minimax-M2.5 struggled to perform orchestration restricted to branch selection. We note again that the orchestrator does not contribute any ideas for subagents to pursue. Its subagent tool only gives a branch field for selecting parent solutions. Therefore, the model configuration faithfully tests whether using a sufficiently powerful orchestrator to guide scaling exceeds fixed scaling. The orchestrator also has a 60 iteration budget and results are averaged over 3 runs.

\subsection{Comparison to Baselines}

\begin{figure}[t]
    \centering

    \begin{minipage}[t]{0.495\linewidth}
        \vspace{0pt}
        \centering
        \includegraphics[width=\linewidth]{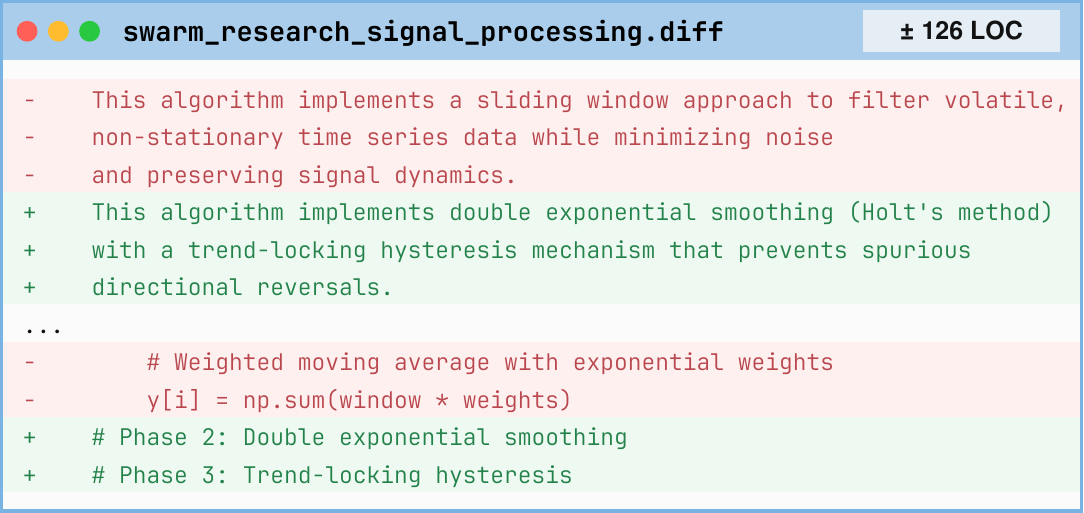}
    \end{minipage}
    \hfill
    \begin{minipage}[t]{0.495\linewidth}
        \vspace{0pt}
        \centering
        \includegraphics[width=\linewidth]{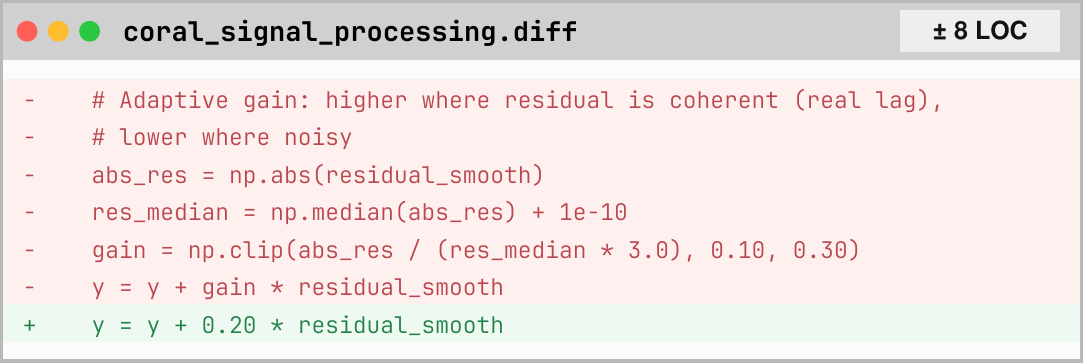}
    \end{minipage}

    \caption{Examples of median-sized \tech and CORAL diffs for \textit{Signal Processing}.}
    \label{fig:example_diffs}
\end{figure}

As shown in Table~\ref{tab:copmarison_table}, \tech exceeds the evolutionary baseline EvoX on 13/15 tasks and matches it on 1 task. \tech strictly exceeds the multi-agent baseline CORAL on 10/15 tasks and matches it on 2 tasks. However, differences between CORAL and \tech on the mathematical optimization tasks Circle Packing, Erdos Min Overlap, and MMD-14-3 are small. 
They could be reduced by further solution polishing under higher budgets. \tech succeeds by exploring diverse high-level approaches while CORAL experiments with diverse but low-level changes on fewer high-level approaches. For example, in Figure~\ref{fig:example_diffs}, \tech's median sized diff involves trying a new type of signal processing algorithm while CORAL makes a targeted change to how signal corrections are computed. As another example, in AHC016, a median sized diff from CORAL involves tuning thresholds while a median size diff from \tech redesigns the solver into different modes applied on different conditions. On most tasks, higher-level experiments mean \tech discovers better solutions. However, within the budget, CORAL better exploited strong approaches and discovered meaningful low-level changes on the systems task EPLB and heuristics task AHC026.

\definecolor{bestgreenbg}{RGB}{215,245,215}
\newcommand{\best}[1]{%
  \begingroup
  \setlength{\fboxsep}{2.5pt}%
  \colorbox{bestgreenbg}{#1}%
  \endgroup
}

\begin{table}[t]
\centering
\caption{Comparison of \tech to state-of-the-art Evolutionary and Multi-Agent Baselines. \emph{We run each approach once with Opus 4.6 and publicly available implementations}. \tech and multi-agent baseline CORAL use Claude Code for \$50 per task. EvoX runs for 100 iterations (\$23.50 per task on average). \textbf{Bolded} means top score among these methods. Highlighted in \best{green} means score matches or exceeds state-of-the-art set by AI systems or humans.}
\begin{tabular}{c|l|cccc}
\toprule
 & \textbf{Task} & \textbf{SOTA} & \textbf{EvoX} & \textbf{CORAL} & \textbf{SwarmResearch} \\
\midrule
\multirow{5}{*}{\rotatebox{90}{\textbf{Math}}}
& Circle-Packing $\uparrow$
    & 2.635983 & 2.1064 & 2.635985 & \best{\textbf{2.635996}} \\
& Signal Processing $\uparrow$
    & 0.8229 & 0.7181 & 0.7403 & \textbf{0.7970} \\
& Erd\H{o}s Min Overlap $\downarrow$
    & 0.380876 & 0.38195 & 0.381099 & \textbf{0.381080} \\
& MMD-14-3 (Min-Max-3) $\downarrow$
    & 4.16578 & 4.46410 & \textbf{4.16578} & 4.16584 \\
& 3rd-Autocorrelation $\downarrow$
    & 1.45368 & 1.46220 & 1.46233 & \textbf{1.45649} \\
\midrule
\multirow{4}{*}{\rotatebox{90}{\textbf{Systems}}}
& EPLB $\uparrow$
    & 0.1490 & 0.1443 & \textbf{0.1467} & 0.1436 \\
& LLM-SQL $\uparrow$
    & 0.7310 & 0.7253 & 0.7195 & \best{\textbf{0.7331}} \\
& Txn Scheduling $\uparrow$
    & 4566.0 & 4081.6 & 4201.7 & \textbf{4366.8} \\
& Cloudcast $\downarrow$
    & 618.4 & 696.1 & \best{\textbf{618.0}} & \best{\textbf{618.0}} \\
& PRISM $\uparrow$
    & \best{\textbf{26.26}} & \best{\textbf{26.26}} & \best{\textbf{26.26}} & \best{\textbf{26.26}} \\
\midrule
\multirow{5}{*}{\rotatebox{90}{\textbf{Heuristics}}}
& Territory (AHC008) $\uparrow$
    & 3463 & 1161 & 1208 & \textbf{1304} \\
& Halloween Candy (AHC015) $\uparrow$
    & 3506 & 1985.67 & 2268 & \textbf{2543} \\
& Graphorean (AHC016) $\uparrow$
    & 3517 & 1621 & 1782 & \textbf{2138} \\
& Balancing by Balance (AHC025) $\uparrow$
    & 3479 & 1400 & 1236 & \textbf{1470} \\
& Stack of Boxes (AHC026) $\uparrow$
    & 3451 & 2009.67 & \textbf{2618} & 1992 \\
\bottomrule
\end{tabular}
\label{tab:copmarison_table}
\end{table}

State-of-the-art results on the math and systems tasks are all claimed by AI systems, though many are difficult to verify because their solutions are not publicly released. On the other hand, the state-of-the-art results on heuristics tasks are all human-authored, and AI systems still lag behind. While math and systems tasks are more widely studied and more standard solution patterns are successful, heuristic tasks are competition problems whose best solutions have been intensively optimized by expert humans using creative, problem-specific approaches. We also reiterate that heuristic task scores indicate competition skill level: a score of 3000 versus 1500 does not imply a 2$\times$ higher objective metric, but corresponds to a higher Elo-like performance level where a 3000-rated competitor would outperform a 1500-rated competitor with high probability. 

\begin{resultbox}
\textbf{\tech's performance exceeds or matches evolutionary and multi-agent baselines on 13/15 open-ended optimization tasks, indicating more effective search of their solution spaces.} \tech exceeds EvoX on 13 tasks while matching it on 1 task. \tech exceeds CORAL on 8 tasks while performing similarly on 5 tasks.
\end{resultbox}

\subsection{Granularity of Code Changes.}
\begin{figure}[h]
    \centering
    \includegraphics[width=0.75 \linewidth]{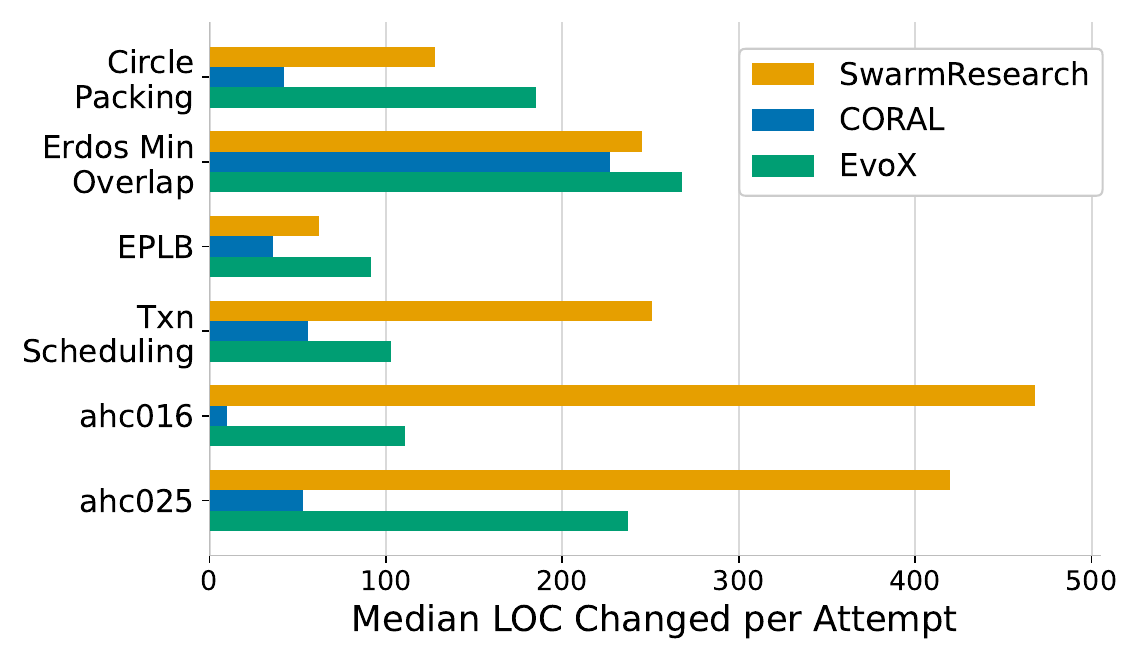}
    \caption{Median lines of code changed per attempt by \tech and baselines.}
    \label{fig:change_granularity}
\end{figure}

As shown in Figure~\ref{fig:change_granularity}, \tech typically makes larger code changes than the baselines. This difference is most prominent in heuristics tasks and some systems tasks like Transaction Scheduling. These tasks have large baseline solutions, and implementing new approaches require more code than other tasks. On mathematical optimization tasks like Circle Packing, experiments involve smaller changes like hyperparameter tuning or optimizer tweaks. Since low-level changes appear as targeted edits while high-level changes in approach appear as larger code changes, this result confirms our context and git management strategies increase the share of experiments that test high-level changes. Although high-level experiments are productive for most tasks, they can be inefficient when they lead nowhere and inference is better spent on optimizing strong approaches.

\begin{resultbox}
\textbf{\tech's experiments test more high-level changes than the baselines.} For the average attempt and median task, \tech changes 3.2$\times$ more lines than CORAL and 1.7$\times$ more than EvoX.
\end{resultbox}

\subsection{Comparison to Fixed Scaling}
\label{section:fixed_scaling}

\begin{table}[h]
\centering
\caption{Comparison of Orchestrator-guided Scaling to Fixed Scaling. The depth of orchestrator-guided scaling is the longest chain of serially spawned search agents, and its width is the average number of parallel agents spawned at each depth.}
\resizebox{\linewidth}{!}{
\begin{tabular}{lcccc}
\toprule
\textbf{Task}
& \textbf{Optimal Fixed Scaling}
& \textbf{Optimal Fixed Scaling}
& \textbf{Orchestrator-guided}
& \textbf{Orchestrator-guided } \\
& \textbf{Width $\times$ Depth}
& \textbf{Score}
& \textbf{Avg. Width $\times$ Depth}
& \textbf{Score} \\
\midrule
3rd-Autocorrelation $\downarrow$
  & 5$\times$12
  & \textbf{1.4649}
  & 8.57$\times$6
  & 1.4652 \\
Signal Processing $\uparrow$
  & 15$\times$4
  & 0.6369
  & 6.78$\times$8
  & \textbf{0.6787} \\
Circle-Packing $\uparrow$
  & 20$\times$3
  & 2.244
  & 9.60$\times$4
  & \textbf{2.294} \\
Cloudcast $\downarrow$
  & 10$\times$6
  & 714.3
  & 10.83$\times$5
  & \textbf{672.3} \\
Txn Scheduling $\uparrow$
  & 20$\times$3
  & 3775
  & 6.00$\times$4
  & \textbf{4049} \\
\bottomrule
\end{tabular}
}
\label{tab:orch-guided}
\end{table}

Table \ref{tab:orch-guided} compares orchestrator-guided scaling to optimal fixed scaling. Experiments uses Minimax-M2.5 as the primary model and only 5 tasks since searching for the optimal scaling configuration is costly (details are in section~\ref{fixed_scaling_configs}). Table \ref{tab:orch-guided} shows that optimal fixed scaling is wider than deeper for 4/5 tasks. However, the most wide configuration $(30,2)$ is never optimal. Overall, we find that shorter parallel runs reach better performance than longer serial runs, up to a point where higher parallelism is not worth it. In longer serial runs, agents successfully refine an approach but stay within the same neighborhood of solutions and miss stronger neighborhoods discovered by highly parallel runs. Additionally, the optimal number of parallel agents and serial iterations under fixed budget varies with task. 

Orchestrator-guided scaling, which requires no scaling hyperparameter selection, exceeds optimal fixed scaling on 4/5 tasks. In fixed scaling, the maximum number of parallel and serial iterations on top of any solution is fixed. However, orchestrator-guided scaling is more flexible. The orchestrator can decide to allocate the iteration budget to wide initial parallel exploration. Then, it can selectively apply deep serial iteration to the most promising solutions. In effect, the agent can express varying width at different depths. For example, orchestrator-guided scaling's max depth is much higher than fixed scaling for \textit{Signal Processing} which allows it to iterate on the best solution for longer. The 60 subagents dominate inference cost, so the orchestrator only increases total output tokens by 7.7\% over subagent-only. 

\begin{resultbox}
\textbf{Orchestrator-guided scaling exceed optimal fixed scaling.} By varying parallelism at different depths, orchestrator-guided scaling achieves better performance than optimal fixed scaling on 2/3 mathematical optimization tasks and 2/2 systems optimization tasks.
\end{resultbox}

\subsection{Shepherd Agent Behavior} 
With minimal prompting, typical Claude Opus 4.6 Shepherd Agent behavior involves launching waves of $\sim$4-8 Search Agents. Waves mostly consist of Explorer Agents, while Optimizers Agents are spawned when dominant solutions emerge. Its default search behavior is near-greedy. The Shepherd Agent begins by spawning concurrent explorers and keeps a few high-level approaches open. Then, it assigns concurrent search agents to iterate on the top-few solutions. Eventually, the Shepherd Agent concentrates Search Agents onto the single, top approach. It very rarely merges agent branches. Its prompts to Search Agents typically prescribe specific ideas to pursue. On plateaus, the Shepherd Agent spawns more concurrent Explorer Agents from the best solution but also from the baseline, with context on the top-performing approach. 

We prompt out negative Shepherd Agent behaviors like assigning specific ideas to Search Agents and collapsing onto one approach, while reinforcing positive behaviors like focusing Search Agents on discovering new approaches from the baseline on plateaus. However, even with thorough prompting, the Shepherd Agent's ability to strategically steer Search Agents without prescribing exact ideas is limited. For example, it struggles to synthesize limitations into prompts or pose specific problems to different Search Agents based on current bottlenecks. Improving an agent's ability to plan strategic experiments is an open problem that would advance discovery ability.

\section{Case Study: Approximate Speculative Decoding}
\label{sec:spec-dec-case-study}

We use speculative decoding as a detailed case study to illustrate \tech's capabilities and limitations. Figure~\ref{fig:evo-graph} visualizes \tech's search trajectory for the task.

\textbf{Task \& Setup.} Speculative decoding techniques are commonly lossless~\citep{leviathan2023fast, hu2025samdecoding, li2025eagle3}, where the target model verifies every draft token to ensure the output exactly matches what the target model would have generated. However, strict per-token verification may not be necessary for preserving downstream task accuracy, and trading some accuracy for higher speed may be desirable~\citep{biglittlespecdec, huang2025specdecpp, lu2025adasd}. To explore this direction with \tech, our fast evaluator includes 30 total reasoning-intensive tasks from LiveCodeBench v6, AIME 2026, GPQA Diamond, and HLE-MCQ~\citep{jain2024livecodebench,matharena2026aime,rein2023gpqa,phan2025hle}. The objective of \tech is to maximize token throughput while preserving benchmark accuracy. The baseline is a basic single-file, lossless speculative decoding implementation using rejection sampling (1.68× faster than vanilla decoding). \tech use Claude Code Opus 4.8. Before solution generation, an initial \tech run analyses the baseline from different angles without generating solutions and its findings are included in the solution generation run worktrees. See details in Appendix~\ref{app:spec-dec}.

\textbf{\tech's Solutions.} On 100 held-out tasks, averaged across 5 seeds, \tech achieves a mean tok/s that is 4.58$\times$ faster than naive vanilla decoding, with 60.6\% accuracy, while vanilla decoding's baseline accuracy is 65.8\%. Vanilla autoresearch~\citep{autoresearch} achieves 1.80$\times$ speedup and CORAL~\citep{coral2026} achieves 2.26$\times$ speedup, with accuracies of 65.8\% and 58.4\% respectively. These methods generated solutions for $\sim$12 hours. For \tech, accuracy loss is due to more responses exceeding our evaluator's 16,384-token limit, despite some reasoning chains including the correct answer. 

The largest throughput gains with minimal accuracy loss come from systems optimizations. The best-performing optimization is batching the target forward pass across multiple benchmark sequences. To preserve accuracy for the evaluated sampling seed, \tech uses independent RNG streams across batches for sampling independence. Autoresearch and CORAL's solutions do not use batching. Figure~\ref{fig:evo-graph} shows \tech's 23rd completed Search Agent applied batching on a fresh branch from main, even after significant serial exploration on approaches like relaxed acceptance conditions. Additionally, \tech's best solution optimizes the construction of token distributions for rejection sampling by using a persistent thread pool, top-$k$ bounded distributions, and parallel distribution construction during batching. The best solution additionally adapts the number of target-model experts. The baseline uses $k=8$ experts throughout, but \tech finds that $k=8$ is only necessary during prefill. For target-token generation, \tech uses $k=4$ for most tokens and increases to $k=5$ when the previous verification round's mean top-1 probability is less than 0.7. Unlike batching and distribution-construction optimizations, adaptive MoE expert count is not strictly lossless with respect to the original target model. 

Existing inference engines~\citep{zheng2024sglang} already implement batching for speculative decoding and optimized token-distribution processing, while adaptive expert count has been explored outside speculative decoding~\citep{balsamo2026adaptivek}. Other than systems optimizations, \tech explores decoding techniques that trade speed and accuracy, including relaxed draft acceptance and skipping verification for confident tokens. Prior work explores related techniques~\citep{biglittlespecdec, huang2025specdecpp, lu2025adasd}. Novel attempts, not in existing decoding literature as far as we know, perform poorly (see risk and challenges for details). 

In one relaxed-acceptance variant, a draft token is accepted directly if: (1) it is also the target model's top-1 token, and (2) either the target   is low (< 1.098), or the target top-1 probability exceeds the target top-2 probability by a large margin ($\geq$0.30). Condition 1 essentially implements greedy speculative decoding. This does not preserve the intent of rejection sampling, but it increases the draft-token acceptance rate. Condition 2 is intended to identify tokens that are less likely to be critical for accuracy, at least based on single-seed observations. This approach is not used by the final solution because it causes accuracy loss when combined with batching. When combined with the final solution and evaluated on the 30-task benchmark across 5 seeds, this approach increases mean tok/s by 9\%, and increases the draft acceptance rate from 76.8\% to 84.8\% while reducing accuracy from 91.3\% to 83.3\%. Most accuracy loss is due to hitting the benchmark's token cap.
 
\textbf{Risks and Challenges} \tech's novel attempts, not in existing decoding literature as far as we know, performed poorly and we did not find them reasonable. For example, one attempt was "comonotone sampling": during rejection sampling for a draft span, it sampled a single shared acceptance threshold for all tokens in the span, rather than sampling independent thresholds per token as is typical. The model produced a run with improved performance and reasoned the change increased the acceptance rate. However, after we generated acceptance rate data and evaluated additional seeds, we found its claims were unsubstantiated. \tech generates a high throughput of ideas and experiments and careful review requires high effort. Without careful review, users may be convinced by low-quality approaches, where proposals and justifications seem attractive at first glance despite being incorrect.

\section{Related Work}

\textbf{LLM-Guided Evolutionary Discovery.} 
A growing number of work are developing \llm-based search and discovery approaches~\citep{romera2024funsearch}.
Initially, the focus has been on using \llm as a mutation operator and combining it with heuristics as part of an evaluator-guided evolutionary loop~\citep{novikov2025alphaevolve, openevolve, lange2025shinka}.
These \llm-guided evolutionary algorithms follow a pipeline process of selecting parents using predefined heuristics based on evaluation feedback and then use an \llm to generate the next round of mutation.
AlphaEvolve~\citep{novikov2025alphaevolve} selects new candidates for future mutation using the MAP-Elites algorithm~\citep{mouret2015mapelites}. 
Follow up work improves search algorithms with techniques like adaptive sampling~\citep{lange2025shinka, openevolve}, pareto frontiers~\citep{agrawal2025gepa}, island-based architectures~\citep{assump2025codeevolve}, and momentum-based backtracking~\citep{yan2026pacevolve}. 
Compare to using fixed candidate selection and variation mechanisms, AdaEvolve~\citep{cemri2026adaevolve} and EvoX~\citep{liu2026evox} also evolve the search strategy itself. However, these works also use LLMs inside a fixed pipeline rather than leveraging their capabilities as open-ended agents.
Additionally, other work focus on training \llm{s} to perform better discovery.
For example ThetaEvolve~\citep{wang2025thetaevolve}, TTT-Discover~\citep{yuksekgonul2026learning}, and FLEX~\citep{cai2025flex} use reinforcement learning to iteratively improve the \llm during the evolutionary loop at test-time.

\textbf{Agentic Solutions for Open-ended Discovery.} 
In addition to LLM-guided evolutionary approaches, other work uses \llm agents to directly carry out open-ended tasks without constraining the \llm to specific operations.
AI Scientist~\citep{lu2024aiscientist} and AI Co-Scientist~\citep{gottweis2025towards}, developed by Sakana AI and Google respectively, are examples of agentic systems that assign \llm agents specialized roles, such as brainstormer and reviewer, to perform scientific discovery.
More recently, autoresearch~\citep{autoresearch} has spurred growing interest in autonomous research agents. It uses a general-purpose coding agent, such as Codex or Claude Code, to iteratively modify research code, run experiments, evaluate results, and keep or discard changes.
Researchers and industry practitioners have developed more advanced solutions with better memory management~\citep{chen2026avo}, shared skills~\citep{hive}, and multi-agent collaboration~\citep{coral2026}. 
Yet, these approaches still use one or more self-organized agents, each in long-running experimental loops with one editable program state. Instead, \tech's orchestrator-subagent architecture avoids context accumulation in a few agents and idea convergence from shared memory (see Section \ref{sec:context_management} for details), while its branch-based versioning preserves competing approaches to avoid greedy local search (see Section \ref{sec:git_management} for details). Additionally, unlike existing multi-agent approaches, which involve complex orchestration such as heartbeat events and require manually configuring the number of agents \tech is fully implemented as skills and requires no hyperparameters since the orchestrator adapts the number of parallel and serial agents autonomously. 

\textbf{Test-time Scaling.} 
Using more LLM inference-time computation for better task performance is known as test-time scaling~\citep{snell2024scaling, brown2024large}. 
A straightforward way to perform test-time scaling is to do \textit{best-of-N} sampling:  sample multiple outputs in parallel from an \llm and select the one that obtains the highest score ~\citep{cobbe2021training, lightman2023let}.
This is also known as parallel test-time scaling.
On the other hand, sequential test-time scaling iteratively prompts the \llm to revise or update their previous output, sometimes based on evaluator feedback~\citep{qu2024recursive}. 
Test-time scaling has been demonstrated to be very effective for challenging domains like software engineering ~\citep{muennighoff2025s1, ehrlich2025codemonkeys, xia2025agentless}. A recent method, SimpleTES~\citep{ye2026evaluation}, achieves impressive results on open-ended discovery by scaling an evaluation-driven workflow across a fixed number of parallel independent trajectories, sequential refinement steps, and best-of-$K$ candidate selection between refinement steps. While existing work uses heuristics or pre-configures scaling, \tech demonstrates how an LLM can autonomously and effectively adapt the degree of parallel and serial scaling on the fly.

\section{Conclusion}
\tech elicits high-level exploration from coding agents through simple harness design choices: improved context management through an orchestrator-subagent architecture and improved program versioning through git branching. Our minimal implementation of these choices as skills makes the approach interoperable with popular coding agent tools. We demonstrate these choices produce a harness that discovers stronger solutions than state-of-the-art evolutionary and multi-agent baselines through higher-level experiments. However, major challenges still remain. Our harness does not enhance the basic creativity of LLMs, so approaches it explores still tend towards variations of known approaches. Additionally, complex Shepherd Agent behavior is prompt-reliant. Models we test do not exhibit sophisticated Search Agent steering behavior, limiting their ability to meet the full potential of the \tech architecture. Our intention is that our minimal harness is usable by practitioners, serves as a base for future harnesses and training on open-ended discovery, and inspires future work on using orchestrators to steer Search Agents.

\bibliography{reference}
\bibliographystyle{colm2026_conference}

\clearpage
\appendix

\begin{center}
    {\Large \bfseries Appendix}
\end{center}

\vspace{1em}

\section{Speculative Decoding Details}
\label{app:spec-dec}
\textbf{Evaluation Details.} The target model is gemma-4-26B-A4B-it, and the draft model is its corresponding gemma-assistant MTP head. All methods have access to and are evaluated on 8xA6000s. Temperature 1.0 is used.

\textbf{Analysis Phase.} We do 2 sequential \tech runs: an $\sim$11-hour analysis phase and an $\sim$11-hour solution generation phase. The goal of analysis it to collect data on the baseline from different angles without generating solutions. For example, Explorer Search Agents profile the baseline and measure target vs. draft confidences. The Shepherd Agent’s role is to maintain a diverse population of analysis while going deep on promising directions. The analysis phase’s findings are organized into a directory accessible to the solution generation phase’s Search Agents. This helps structure Search Agent idea generation, so its ideas are more novel but motivated by specific observations. As an example, analysis finds that most of the target model's generation consists of low-entropy tokens that the draft predicts well, but there are few high-entropy tokens that the draft model does not predict well and throws of the reasoning trace if incorrect.

\section{Between-run Variance}
\label{app:variance}
Between-run variance is a special challenge for discovery technique reproducibility. Discovery techniques rely on high-temperature sampling over uncertain LLM token distributions to open-ended queries, which drives the apparent creativity of LLMs. Given that LLM variance forms a basis of many discovery techniques, some variance between independent runs can be hard to avoid. To achieve a stable estimate of some discovery techniques e.g. vanilla autoresearch~\citep{autoresearch}, a researcher might have to run it over 5 times for some tasks. Even more so, we have have observed that the number of independent runs needed to achieve a stable estimate can be task-specific. However, running discovery techniques multiple times is infeasibly expensive when using frontier models, which are important to use since they discover better solutions than cheaper models and they exhibit unique behaviors a technique might rely on. For instance, just one run of all three approaches on our 15 task benchmark cost $\sim$\$1700 worth of Opus 4.6 credits. 

In light of the costliness of these techniques, the number of independent runs should be part of the recommended use of a discovery technique. This recommendation is crucial for both users and researchers to know how to allocate their budget to independent runs when using and reproducing a technique. For example, given a \$50 budget of frontier model credits per task, a technique should inform users whether splitting it into 3 runs of \$17 or doing 1 long \$50 run will achieve the most representative results. Exact reproducibility can be challenging for some discovery techniques, but these guidelines inform users and researchers on how to best achieve a \textit{representative} result for both real-world use and benchmarking.

For our approach, we recommend one run since independent search agents are a form of parallel scaling that improve between-run stability. We also run the baseline approaches once since independent runs are not in the baselines’ recommendation for users. This way, under realistic budget constraints, we can reproduce the most representative result for the baselines. Although this means that our reported baseline results might not reflect a stable estimate across many independent runs, we view this as the best option given their stochastic and costly nature.

\section{Skills}
\label{app:skills}

\subsection{Shepherd}
\begin{skillprompt}
---
name: shepherd
description: How to organize a population of search agents
---

You organize a population of search agents towards multiple breakthroughs to an open-ended research problem. Your goal is to manage the population to follow these invariants:
- Initialize a diverse population of ideas 
- Maintain a diverse population of ideas and avoid idea collapse throughout the discovery process
- Prioritize effort on promising ideas
- Make consistent progress and break out of plateaus

You accomplish these goals by spawning patterns of *explorer* and *optimizer* search agents, selecting their parents, and adding minimal context as needed. *The most important rule*: search agents independently determine their experiments; you are not allowed to instruct them to work on specific ideas. 

Keep going until you've exhausted your explorer budget. Your explorer agent budget is *25*. Keep count of how many you've spawned. Don't give up, keep searching for new families or launching serial agents.

### Search agent types
- *Explorer*: Use to increase diversity of solutions and to break out of plateaus. Increase the number of concurrent explorers to increase diversity, even from the same parent or from different parents for even greater diversity. Explorers will experiment with one new idea and return a summary.
- *Optimizer*: Use to optimize best solutions. Only one optimizer from a given parent solution should be alive at a time. Optimizers experiment with a few successive edits and tweaks to improve a solution. 
- How they work:
  - All search agents frequently commit their recent solutions to their assigned branch. Their commit messages include evaluated scores.
  - *Serial search agents* (search agents spawned one after another as each other's parent solution) remember work done by their ancestors, so they will try new ways to make progress along a similar direction. 

### Adapting search agent patterns
You should constantly think and reflect on current progress to select the right search agent pattern. Adapt the type of search agent, size of fan-outs, amount of serialization, etc. to current progress. Some examples of patterns:
- *Explorer Fan-out*: Multiple explorers from one parent. Use on plateaus to explore multiple promising parents. More concurrent explorers increases diversity. However, emphasize serializing explorer agents which strategically build on top of sucesses and failures while concurrent explorers are unaware of each other. 
- *Serial Explorers*: Select promising explorer parents from prior wave. Launch more explorers using them as parents. Actuate the number of explorers in each fan-out. 1 can be enough. You can add more after they're complete if not enough.
- *Multi-parent*: Select multiple parents for an agent so they can combine or take inspiration from them
- *Serial optimization*: Optimizers spawned one after another. Many successive optimizers make slow but eventual progress.
- *Combinations*: Interleave serial explorers, optimizers, fan-outs.

**Be creative. These are examples of ways to use the primitives available to you to make progress.**

### Searching for better families of approaches from initialization and on plateaus
1) **A substantial portion of your explorer budget should go towards building a large, diverse initial population** of ideas from master. Then, continue to spawn serial explorers ontop of promising diverse parents. Keep going until exploration plateaus.
- Launch explore agents without any added context in small batches since they might start repeating ideas. 
- If explore agents collapse onto similar ideas, nudge exploration by providing a summary of already explored ideas and prompt to explore new ideas as context. NEVER tell them which ideas to explore, just what not to explore again.
- During this stage, do NOT spawn optimizers. Optimizers are launched late once there are some clear bests.
- Wait until explorers are complete before much serial development, long-running explorers might have the best solutions.
2) **Spawn long sequences of serial explorers and optimizers** for multiple promising families.
Spawning serial agents on solutions that didn't improve is also productive. Serial search agents remember work done by their ancestors, so they might make progress along similar directions. However, adapt search agent patterns if this isn't working. Do NOT collapse diversity by spawning serial agents on only one family of ideas. 
3) **Every family eventually hits plateaus, so the population must always develop new families of approaches**. In addition to adding context to explorers on ideas not to repeat, you might add minimal context on strengths and weakensses of current approaches to support strategic ideas on plateaus.

### Add minimal context as needed
Search agents only have knowledge of their branch. They do NOT share your your context and do NOT have context on other search agents. Only add context if necessary and keep it minimal and non-prescriptive. You might:
  - Provide a summary of already explored ideas and prompt them to explore new, more creative ideas. 
  - On plateaus, provide a 1) briefing of strengths and weaknesses of existing families and 2) a suggestion to identify a novel technique that solves the limitations
  - Add context of relevant findings from other search agents that are helpful to refine their solution. Only add complementary findings. 
  - Control the level of explorativenss. If the agent should try radical ideas, then say so. If it should explore smaller refinments, then say so. 

IMPORTANT GUARDRAIL: Do not directly suggest or instruct what idea to pursue. Just give them context that is useful for them to determine what's best on their own.
- GOOD example of cross-agent context: "Your parent scored 0.78 with strong final answer quality and speed, but it still violates hard constraints in edge cases. Another branch reached 0.79 through a different family that improved constraint satisfaction and consistency, but it was less accurate on ordinary cases. Use those observed tradeoffs as context when deciding what experiment to run."
- BAD example of context: "Here is a completely fresh angle that no one has tried: build a beam-search planner with a learned scoring model and tune beam width in [8, 16, 32]." This is bad because it assigns a specific method and parameter direction.
- BETTER version used on plateau: "The current leading family is strong on final answer quality but still fails on hard constraint satisfaction. Several constraint-first alternatives improved validity but lost too much answer quality. Look for a different tradeoff that addresses this failure mode without assuming the current repair step is the right mechanism."

### Selecting parents by creating new branches
The neighborhood explored by search agents depends on their parent solution which serves as their solution's starting point. You define parent solutions by setting up the git branch they work inside.
- Spawn search agnets on diverse parents. You don't know squat, so never collapse onto a single idea.
- Don't end ideas prematurely. If a search agent encounters a promusing idea that doesn't beat an idea you've already spent resources developing, consider refining and exploring it further. It might be a new neighborhood containing an improved solution.

Every agent must receive a branch that already exists. Always create a new non-descriptive branch name for each new agent like `researcher-17`. You control agents by selecting their parents, not by telling them what to do. 
- Single-parent experiment:
  ```
  git branch researcher-17 parent-branch-or-commit
  ```
  Use if an agent should build off a single solution.
- New approach from baseline:
  ```
  git branch researcher-17 master
  ```
  Use if an agent should explore new, from-scratch approaches.
  Use `main` instead of `master` if this repository uses `main`.
- Multi-parent experiment:
  ```
  mkdir -p .worktrees
  git worktree add -b researcher-17 .worktrees/researcher-17 first-parent
  cd .worktrees/researcher-17
  git merge --no-commit second-parent
  ```
  Use if an agent should take inspiration from multiple solutions or combine them.
  Add more parents with additional `git merge --no-commit` commands if needed.
  Merge conflicts are expected. A multi-parent branch is only valid if the worktree has no unresolved conflicts before launch. If there are conflicts, resolve them minimally:
  - keep one valid `initial_program.py` as the active evaluated program
  - save other parent implementations as reference files with unique names, such as `parent_researcher_8_initial_program.py`
  - combine or keep `findings.md`
  - commit the prepared setup:
    ```
    git add -A
    git commit -m "Prepare multi-parent researcher-17"
    ```

Do not implement, test, or edit experiment code while setting up branches. Branch setup only.

### Spawning agents
Spawn as many search agents as needed using the following commands. Follow their format exactly with every flag included. Include context in session prompts as needed. Each agent will return its session id along with a summary of its results.
- To spawn an `explorer`, start a fresh agent session:
```
claude --permission-mode bypassPermissions -p "Follow the explorer skill. Your branch name is branch_name. <optional: context>" --output-format json | jq -r '"session_id=\(.session_id)\nmessage=\(.result)"'
```
- To spawn an `optimizer`, resume and fork an existing agent session (choose just one session if merging):
```
claude --permission-mode bypassPermissions --resume <parent_session_id> --fork-session -p "Follow the optimizer skill. Your new branch is <branch_name>. <if applicable: note that additional findings were merged in> <optional: context>" --output-format json | jq -r '"session_id=\(.session_id)\nmessage=\(.result)"'
```
- Your handoff prompt to agents must follow the exact given format. Do NOT give it any specific directions or suggestions. ONLY give it the branch name you setup for it and minimal context.
- Stay in one directory. Switching directories causes claude to discover the session in a different namespace than the one it was originally spawned in, which causes search agents to fail to spawn.
- Background agent completion is reported by task notifications. Wait for those notifications; use coarse polling only as a fallback. Agents can take a long time.

**Keep spawning new agents until the budget is exhausted.** Progress is always possible by spawning more serial agents and finding a better family of approach. ALWAYS create new branches for every agent.
\end{skillprompt}

\subsection{Explorer}
\begin{skillprompt}
---
name: explorer
description: Read if asked to read the explorer skill
---

You are a researcher working on an open research problem. The user might prompt you with additional context or instructions. It's important you take them into consideration. If you miss any step or do not follow its rules, your response will be rejected.

**STEP 1: Create your assigned git worktree in `.worktrees/`**
- Create a new git worktree for your assigned branch at `.worktrees/<assigned-branch>`.
- This worktree is your only working directory.
- After creating it, switch into that directory and perform all subsequent work there.
- Use only `git worktree add .worktrees/<assigned-branch> <assigned-branch>`. If it fails, stop.

IMPORTANT GUARDRAIL: Only use files already in your assigned worktree. Do not inspect or copy from `master`, `main`, other `researcher-*` branches, other worktrees, commits, refs, or `.git`. Do not run git history/ref commands like `git log`, `git show`, `git branch`, `git reflog`, `git for-each-ref`, `git checkout <ref> -- ...`, or `git restore --source <ref> ...`.

**STEP 2: Read prompt.md for the problem description, findings.md (if it exists), and any initial programs in your branch**
Pause and reflect on existing findings, if any. Come up with a fundamentally new idea for an improved solution. Do NOT make small, incremental tweaks. 

**STEP 3: Implement and evaluate one idea**
- Focus on one idea. Do not branch out to other major ideas, stay focused on implementing a single idea. There will be future chances to iterate. 
- Use the standard evaluator for evaluating your solution
- After *every* evaluation, EXPLICITLY DECIDE whether to continue iterating or finish up. We have limited budget so be conservative with your iterations. If you have fully implemented a single idea, then finish up. Don't branch out into other ideas that majorly deviate from the initial idea. If the result suggests a targeted fix or improvement, you may continue editing initial_program.py and run ./task-eval again. If the idea doesn't work after a few attempts, give up immediately. Finish up by creating/editing findings.md as needed and committing your changes."

**Step 4: Create/edit findings.md as needed**
- `findings.md` concisely records failures and successes. It is strictly factual and neutral. 
- Only make edits if you learned new facts. Otherwise, keep as is.

IMPORTANT GUARDRAIL: Do NOT discuss next steps or suggestions on what is necessary to improve the score. Simply record the basic facts of your solution and its evaluation. 

**Step 5: Commit all your changes**
- Commit message should follow this format: `{concise experiment description} | score = {score from standard evaluator or concise failure reason}`. 
  - Do not include a score on failure. Only include a description of the failure reason.

When returning a summary back to the orchestrator, include your commit emssage and the commit hash.
\end{skillprompt}

\subsubsection{Optimizer}
\begin{skillprompt}
---
name: optimizer
description: Read if asked to read the optimizer skill
---

You are improving a solution to an open-ended problem.

IMPORTANT GUARDRAIL: Only use files already in your assigned worktree. Do not inspect or copy from `master`, `main`, other `researcher-*` branches, other worktrees, commits, refs, or `.git`. Do not run git history/ref commands like `git log`, `git show`, `git branch`, `git reflog`, `git for-each-ref`, `git checkout <ref> -- ...`, or `git restore --source <ref> ...`.

## Setup
**Create your assigned git worktree in `.worktrees/`**
- Create a new git worktree for your assigned branch at `.worktrees/<assigned-branch>`.
- This worktree is your only working directory.
- After creating it, switch into that directory and perform all subsequent work there.
- Use only `git worktree add .worktrees/<assigned-branch> <assigned-branch>`. If it fails, stop.

**Read prompt.md for problems description, findings.md (if it exists), and any initial programs in your branch**

Loop for 5 iterations. Keep count.

## Loop 
1) Reflect and think of a clear idea on how to improve the solution

2) Edit the solution

3) git commit

4) Run standard evaluator

5) Amend a commit message
- Commit message should follow this format: `{concise experiment description} | score = {score from standard evaluator or concise failure reason}`. 
  - Do not include a score on failure. Only include a description of the failure reason.

6) If score improved, keep the commit. If score is equal or worse, git reset back to where you started

7) Edit findings.md if needed
- `findings.md` concisely records failures and successes. It is strictly factual and neutral. 
- Only make edits if you learned new facts. Otherwise, keep as is.

IMPORTANT GUARDRAIL: Do NOT discuss next steps or suggestions on what is necessary to improve the score. Simply record the basic facts of your solution and its evaluation. 

**Do NOT stop until you've completed 5 iterations**
\end{skillprompt}

\end{document}